# Adversarial Defense via Neural Oscillation inspired Gradient Masking

Chunming Jiang
University of Canterbury
New Zealand
cji39@uclive.ac.nz

Yilei Zhang*
University of Canterbury
New Zealand
yilei.zhang@canterbury.ac.nz

**Abstract**

Spiking neural networks (SNNs) attract great attention due to their low power consumption, low latency, and biological plausibility. As they are widely deployed in neuromorphic devices for low-power brain-inspired computing, security issues become increasingly important. However, compared to deep neural networks (DNNs), SNNs currently lack specifically designed defense methods against adversarial attacks. Inspired by neural membrane potential oscillation, we propose a novel neural model that incorporates the bio-inspired oscillation mechanism to enhance the security of SNNs. Our experiments show that SNNs with neural oscillation neurons have better resistance to adversarial attacks than ordinary SNNs with LIF neurons on kinds of architectures and datasets. Furthermore, we propose a defense method that changes model's gradients by replacing the form of oscillation, which hides the original training gradients and confuses the attacker into using gradients of 'fake' neurons to generate invalid adversarial samples. Our experiments suggest that the proposed defense method can effectively resist both single-step and iterative attacks with comparable defense effectiveness and much less computational costs than adversarial training methods on DNNs. To the best of our knowledge, this is the first work that establishes adversarial defense through masking surrogate gradients on SNNs.

## 1. Introduction

Spiking neural networks (SNNs) recently attracted more and more attention due to their biological plausibility. In addition to neurons and synapses, SNNs incorporate the concept of time into models. Neurons in SNN receive spike trains as inputs, and these spike trains will increase or decrease their membrane potentials. Unlike conventional artificial neural networks (ANNs), the neurons of SNNs transmit information only if their membrane potential reaches a specific firing threshold. Information is sent to the next-layer neurons in the form of spike trains. These characteristics may underline the information transmission and processing in the brain. It is therefore regarded as the next-generation neural network (Tavanaei et al. 2019). Like the brain working fast and efficiently, SNN is also proved to have much better power efficiency (Mostafa et al. 2017) and shorter latency (Wu et al. 2019) compared with ANNs. Besides, researchers also noticed their promising capability in processing dynamic and noisy information (Maass 2014, Deng et al. 2020, Liang et al. 2021). SNNs have been applied in various tasks such as spike pattern recognition (Wu et al. 2019), optical flow estimation (Haessig et al. 2018), and sparse representation (Shi et al. 2017). Since SNNs are being widely deployed in neuromorphic devices such as IBM TrueNorth (Merolla et al. 2014) and Intel Loihi (Merolla et al. 2014), the security aspect of SNNs becomes vital.

In ANNs, models are vulnerable to adversarial attacks that deceive the model into producing the wrong outputs by adding imperceptible perturbations into the clean input. This results in ANNs having catastrophic consequences in certain tasks, such as medical diagnosis and self-driving cars. These attacks are most based on gradients to generate perturbations, such as Fast Gradient Sign Method (FGSM) (Goodfellow, Shlens and Szegedy 2014), Basic Iterative Method (BIM) (Kurakin, Goodfellow and Bengio 2016), and Projected Gradient Descent (PGD) (Madry et al. 2017). Therefore, it is important to improve the robustness of the model and resist the aforementioned adversarial attacks. Several adversarial defense methods were proposed, such as ensemble training (Tramèr et al. 2017), denoising (Xu, Evans and Qi 2017), and adversarial training (Madry et al. 2017).

Despite its popularity in ANNs, adversarial attacks rarely receive any attention in the SNN domain. One reason may be the non-differentiability of spiking events, making supervised learning of SNNs difficult. Some relevant studies of adversarial attacks on SNNs concentrate on gradient-free attack methods (e.g., trial-and-error input perturbation (Bagheri, Simeone and Rajendran 2018, Marchisio et al. 2020)) or spatial gradient-based ANN-to-SNN conversion methods (Sharmin et al. 2019). The computational complexity of the former methods is relatively high due to the absence of the gradient's guidance. The latter lacks temporal components, which leads to inefficient attacks (Liang et al. 2021). Recently, a supervised learning algorithm using a surrogate function to approximate the derivative of spike activity (Wu et al. 2018, Neftci, Mostafa and Zenke 2019, Wu et al. 2019) exhibited success in training high-performance SNNs and raised the opportunity to realize spatiotemporal adversarial attacks on SNNs based on gradients (Sharmin et al. 2020).

*corresponding author

Adversarial defense against adversarial attacks is still in its initial stage for SNNs. There are few literatures devoted to adversarial defense methods for SNNs. In (El-Allami et al. 2021), the authors demonstrate that the simulation time and threshold of SNNs impact the robustness to imperceptible perturbations. However, they do not propose a defense method to resist the interference of adversarial samples effectively. In this work, we propose a specific adversarial defense method for SNN based on a novel bio-inspired approach, where neural oscillation is harnessed for the first time to enhance performances of SNNs under adversarial attacks significantly. We first present a neural oscillation neuron model to train models. The gradients of models will be masked by an alternative neural oscillation after training, thus creating interference in the gradient-based generation of the adversarial samples and effectively enhancing the robustness of the SNNs. We have verified the effectiveness of our defense method on CIFAR-10 and CIFAR-100 datasets (Krizhevsky and Hinton 2009).

In summary, our main contributions are:

1. We propose a novel neural oscillation neuron that is bio-plausible and robust. It blurs the gradients of the SNNs model and interferes with the effect of perturbations on SNNs.

2. We derive an alternative neural oscillation neuron through the neural oscillation neuron. The neuron, being very 'weak', is able to attenuate the attack capability of adversarial samples, thus indirectly enhancing the robustness of the network.

3. Based on two types of neurons, we propose a defense strategy that uses the 'fake' neuron to confuse the attacker and thus achieve adversarial defense. The developed defense method can effectively resist kinds of adversarial attacks, such as FGSM and PGD.

The rest of this paper is organized as follows. Section 2 provides some preliminaries of SNNs and adversarial attacks. The experimental setup and our neural oscillation models are discussed in Section 3. Section 4 validates the validation of our defense methodology. Section 5 concludes this article.

## 2. Preliminaries

### 2.1. SNNs and biological neural oscillation

SNNs adopt spike trains as information carriers between neurons. Every spiking neuron in a SNN receives and emits spikes. The LIF neuron model is a popular bio-inspired simplified model for describing the dynamics of spiking neurons. The dynamics of the LIF model are defined (Fang et al. 2020) by

$$H(t) = \lambda * V(t-1) + \sum_i w_i x_i(t) \quad (1)$$

$$S(t) = \begin{cases} 1, & H(t) > V_{th} \\ 0, & H(t) \leq V_{th} \end{cases} \quad (2)$$

$$V(t) = H(t)(1 - S(t)) + V_{reset} * S(t) \quad (3)$$

where $H(t)$ and $V(t)$ represent the membrane potentials before and after triggering a spike at time $t$, respectively. $V_{th}$ denotes the firing threshold, which is 1 in this paper. $V_{reset}$ is the resting potential, which is 0. $S_t$ denotes the output of neurons at time $t$, $w_i x_i(t)$ is the $i$-th weighted pre-synaptic input at time $t$, and $\lambda$ is the decaying time constant, which is 0.5.

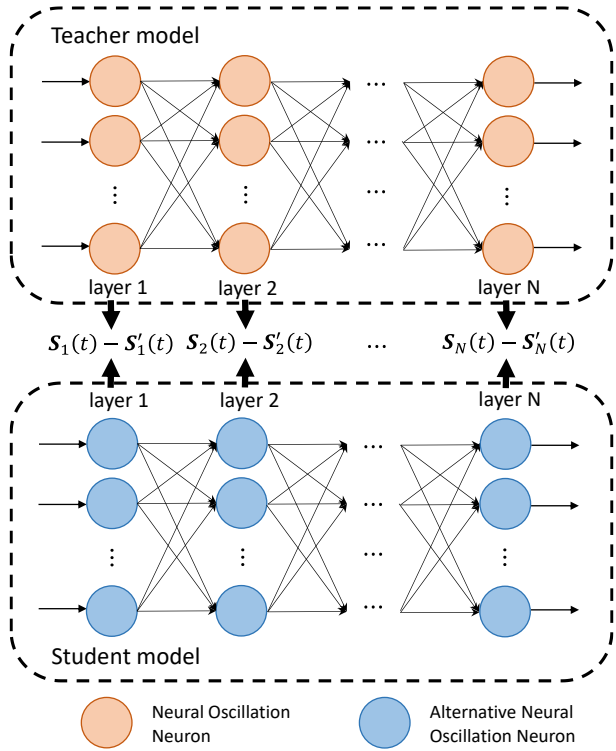

Fig. 1. The training process of the model with alternative neural oscillation neurons at time $t$. The model trained first with neural oscillation neurons can be regarded as a 'teacher model'. It provides the labels for a 'student model' which replaces neural oscillation neurons with alternative neural oscillation neurons. The 'student model' keeps the same trained weights and fits spike trains $S'_j(t)$ of student model to $S_j(t)$ of teacher model by learning variables $a$ and $b$ in each layer.

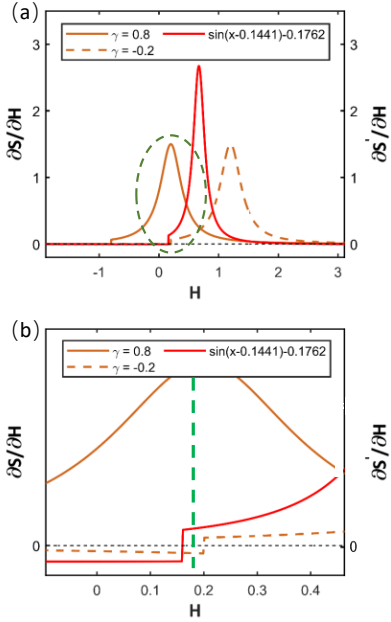

Fig. 2. (a). The solid and dotted orange lines represent $\frac{\partial S}{\partial H}$ of neural oscillation model when $\gamma$ is -0.2 and 0.8, respectively. The red line is $\frac{\partial S'}{\partial H}$ of alternative neural oscillation model. (b). Partial enlargement of graph (a) in the green dashed circle.

According to (1)-(3), when a neuron receives spikes from the previous-layer neurons, its membrane potential will increase. Once the potential value surpasses the neuron's firing threshold, the neuron will fire one spike and promptly be reset to the initial potential $V_{\text{reset}}$.

The biological nervous system generates rhythmic patterns of activity called neural oscillation (Başar 2013). Neural oscillations are thought to be associated with many cognitive functions such as information transfer, perception, motor control, and memory. Such oscillation is mainly triggered by the interaction of individual neurons. In individual neurons, neural oscillation can manifest as the oscillation of membrane potentials or as rhythmic action potentials. This kind of spontaneous activity plays an important role during brain development, including synaptogenesis and network formation. Even though neural oscillations are ubiquitous in biological neurons, the current common spiking neuron models for deep learning, such as IF and LIF models, do not include this oscillatory mechanism, and there is no literature that develops adversarial defense of deep learning models using neural oscillation.

## 2.2. Adversarial attacks

Adversarial attacks (Szegedy et al. 2013) introduce imperceptible perturbation into the input data to mislead the model's classification result. Adversarial attacks can be classified as targeted and non-target attacks according to adversarial goals. A targeted attack is when the attacker attempts to misdirect the model to a class that is different from the true class, while a non-target attack means that the attacker attempts to mislead the model by predicting any of the incorrect classes (Rathore et al. 2020).

In gradient-based adversarial attacks, for a clean image $x$ belonging to class $k$ and a trained SNN model $M$, the adversarial image $x_{adv}$ of $x$ needs to satisfy the following two criteria:

1). The difference between $x_{adv}$ and $x$ is imperceptible, i.e., $\|x_{adv} - x\|_p \leq \epsilon$

2). The model misclassifies $x_{adv}$, i.e., $M(x_{adv}) \neq k$

where the distance metric $\|.\|_p$ denotes the $p$-norm quantifying the similarity, and $\epsilon$ reflects the maximum allowable perturbation on the image.

There are various kinds of adversarial attack algorithms that generate adversarial samples to deceive the model. In this work, we adopt four typical adversarial attacks to evaluate our defense model.

**Fast Gradient Sign Method (FGSM)** (Goodfellow, Shlens and Szegedy 2014) is the most basic approach for generating adversarial samples, which aims at finding a perturbation that maximizes its cost function for the perturbed input (Sharmin et al. 2020). This approach generates adversarial samples by perturbing once the clean image $x$ by the amount of $\epsilon$ along the input gradient direction:

$$x_{adv} = x + \epsilon \cdot \text{sign}(\nabla_x \mathcal{L}(x, y)) \quad (4)$$

Here, $\mathcal{L}$ represents the cost function of the model, and $\nabla_x(*)$ is the model's gradient with respect to a clean sample of $x$. $y$ is the label corresponding to $x$.

**Basic Iterative Method (BIM)** (Kurakin, Goodfellow and Bengio 2016) is an iterative version of FGSM and generates the adversarial samples as:

$$x_m = \text{clip}_\epsilon \left( x_{m-1} + \frac{\epsilon}{i} \cdot \text{sign}\left(\nabla_{x_{m-1}}(\mathcal{L}(x_{m-1}, y))\right)\right) \quad (5)$$

where $x_0$ is the clean image, $x_m$ is an adversarial sample in the $m$-th iteration, and $i$ is the iteration number. $\text{clip}_\epsilon(*)$ represents element-wise clipping of the argument to the range $[x - \epsilon, x + \epsilon]$.

**Momentum Iterative Method (MIM)** (Dong et al. 2018) is similar to BIM but is extended to promote the stability of gradient direction through the addition of a momentum term:

$$g_m = \mu \cdot g_{m-1} + \frac{\nabla_{x_{m-1}}\mathcal{L}(x_{m-1}, y)}{\|\nabla_{x_{m-1}}(\mathcal{L}(x_{m-1}, y))\|_1} \quad (6)$$

$$x_m = \text{clip}_\epsilon \left( x_{m-1} + \frac{\epsilon}{i} \cdot \text{sign}(g_m) \right) \quad (7)$$

$\mu$ is the decaying factor.

**Projected Gradient Descent (PGD)** is one of the strongest iterative adversary attacks. It starts from a random position in the clean image neighborhood $\mathcal{U}(x, \epsilon)$. Its expression is described as:

$$x_m = \text{clip}_\epsilon \left( x_{m-1} + \gamma \cdot \text{sign}\left( \nabla_{x_{m-1}} \mathcal{L}(x_{m-1}, y) \right) \right) \quad (8)$$

where $m$ is the iterative number, and $\gamma$ is the step size.

## 3. Experiments

### 3.1. Datasets and Models

We conduct the experiments on SNN versions of VGG-16 and ResNet-18 for CIFAR-10 and ResNet-18 for CIFAR-100. All models were trained by surrogate gradient-based BP with maxpool layers replaced by average pooling. Bias terms are not included in SNNs. After the convolution layer, we add a batch normalization layer to change the input distribution. Before the fully connected layer, a dropout layer with the probability of P = 0.5 is used to prevent overfitting.

For both CIFAR-10 and CIFAR-100 datasets, all data are normalized to [0,1]. SNNs are trained for 100 epochs with cross-entropy loss and Adam (Kingma and Ba 2014) optimizer. The initial learning rate is set to 1e-4, and the cosine annealing (Loshchilov and Hutter 2016) learning rate schedule with $T_{max} = 100$ adjusts the learning rate over training. A total of 8 timesteps are used for all SNNs. We measure the attack success rate of adversarial sample crafting on 1000 samples randomly selected from each dataset.

Due to the discontinuity of the spiking activity, when training the model, we use the derivative of the Atan function as the surrogate gradient function (see equation (9), $\alpha = 3$) to approximate the derivative of spiking activities.

$$y(x) = \frac{\alpha}{2\left(1 + \left(\frac{\pi}{2}\alpha(x - V_{th})\right)^2\right)} \quad (9)$$

### 3.2. Neural oscillation neuron

Inspired by the biological neural oscillation, we add random oscillation noise in the LIF neuron. We refer to the new neuron as the neural oscillation model. Its dynamic can be described by (1) and (10)-(13).

$$P(t) = f(H(t) + \gamma(t)) \quad (10)$$

$$S(t) = \begin{cases} 1, & P(t) > V_{th} \\ 0, & P(t) \leq V_{th} \end{cases} \quad (11)$$

$$V_t = P(t)(1 - S(t)) + V_{\text{reset}} S(t) \quad (12)$$

$$f(x) = \begin{cases} -0.03x, & x < 0 \\ x, & x \geq 0 \end{cases} \quad (13)$$

$\gamma(t)$ is an independent uniformly-distributed random noise in a range of $[a, b]$ for neurons in each layer, which is [-0.2,0.8] in this paper. $f(x)$ is a piece-wise linear function $LeakyReLU$ whose gradients are defined as -0.03 if $x \leq 0$ and gradients are 1 if $x > 0$.

### 3.3. Alternative Neural oscillation neuron

We train and then save SNN with neural oscillation model, then we copy the model and replace the neural oscillation neuron with a new neural model called alternative neural oscillation. The new neural model changes the noise item $\gamma_i(t)$ in equation (10) to a Sine function of the membrane potential $H(t)$, as (14) describes. The firing and reset actions keep the same as (11) and (12).

$$P(t) = f(H(t) + \sin(H(t) + c) + d) \quad (14)$$

Variables $c$ and $d$ are learnable parameters shared by all layers in the network. The mapping function $\sin(H(t) + c) + d$ is selected here to fit the noise in neural oscillation (More details of the mapping function selection can be found in the supplementary). We freeze all weights of the model with alternative neural oscillation and only keep parameters $c$ and $d$ learnable. The model as the student model was trained again to learn each layer's output of the saved model with neural oscillation neurons, which is regarded as the teacher's model. Here we define the loss function (15) to minimize the difference between the spike trains $S_j(t)$ and $S'_j(t)$ between neural oscillation and alternative neural oscillation in $j$-th layer at time $t$.

$$L = \sum_t \sum_j \frac{1}{2} (S_j(t) - S'_j(t))^2 \quad (15)$$

In this way, the accuracy of models changes slightly (see Table 1). In experiments this process requires only 1-3 training epochs, besides, weight parameters are frozen and few parameters are learnable, thus adding almost no additional training time. The details of noise range $[a, b]$ and values of $c$ and $d$ obtained by training can be found in the supplementary. The process of generating alternative neural oscillation neurons is presented in Figure 1.

### 3.4. Adversarial defense strategy

Now we have two models with different neurons. The two models have approximate output and inference accuracy but with different gradients. For neural oscillation model, we calculate the gradients $\frac{\partial S(t)}{\partial H(t)}$, when $H(t) + \gamma_i(t) \geq 0$,

$$\frac{\partial S(t)}{\partial H(t)} = \frac{\partial S(t)}{\partial P(t)} \frac{\partial P(t)}{\partial H(t)} \approx \frac{\alpha}{2\left(1 + \left(\frac{\pi}{2}\alpha(H(t) + \gamma_i(t) - V_{th})\right)^2\right)} \quad (16)$$

when $H(t) + \gamma_i(t) < 0$,

$$\frac{\partial S(t)}{\partial H(t)} = \frac{\partial S(t)}{\partial P(t)}\frac{\partial P(t)}{\partial H(t)} \approx \frac{-0.03\alpha}{2\left(1+\left(\frac{\pi}{2}\alpha(H(t)+\gamma_i(t)-V_{th})\right)^2\right)} \quad (17)$$

For alternative neural oscillation model, when $H(t) + \sin(H(t)+a) + b \geq 0$, the gradients $\frac{\partial S'(t)}{\partial H(t)}$ is

$$\frac{\partial S'(t)}{\partial H(t)} = \frac{\partial S'(t)}{\partial P(t)}\frac{\partial P(t)}{\partial H(t)} \approx \frac{\alpha(1+\cos(H(t)+a))}{2\left(1+\left(\frac{\pi}{2}\alpha(H(t)+\sin(H(t)+a)+b-V_{th})\right)^2\right)} \quad (18)$$

when $H(t) + \sin(H(t)+a) + b < 0$,

$$\frac{\partial S'(t)}{\partial H(t)} = \frac{\partial S'(t)}{\partial P(t)}\frac{\partial P(t)}{\partial H(t)} = \frac{-0.03\alpha(1+\cos(H(t)+a))}{2\left(1+\left(\frac{\pi}{2}\alpha(H(t)+\sin(H(t)+a)+b-V_{th})\right)^2\right)} \quad (19)$$

We plot both models' gradient distributions of VGG16 for CIFAR-10. As we see, the gradient distribution using alternative neural oscillation is distinguished from the gradient distribution using neural oscillation. The gradient distribution of the alternative neural oscillation has the bigger amplitude and the sharper shape than that in Figure 2(a). On the other hand, in Figure 2(b), we zoom in on the part of the dashed circle in Figure 2(a) and observe the gradient, and it could be seen that gradients vary greatly at the same H value between neural oscillation neuron and alternative neural oscillation neuron.

Our defense strategy is to disguise and hide the real neurons in the model, which confuses the attacker to generate attack samples and attack with the gradients of the 'fake' neurons, which deviate from the real gradients and thus reduce the efficiency of the attack. We validate

**Table 1** Top-1 Accuracy (%) on clean images using two kinds of oscillation neurons

| Model/Dataset | Neural oscillation | Alternative neural oscillation | Accuracy loss |
|---|---|---|---|
| VGG-16/CIFAR-10 | 88.59 | 87.38 | 1.21 |
| ResNet-18/CIFAR-10 | 92.59 | 92.35 | 0.24 |
| ResNet-18/CIFAR-100 | 67.58 | 67.05 | 0.53 |

**Table 2** Neuron model summary under different attack scenarios

| Scenario | Attackers know the real neuron model | Neuron model chosen by attackers to generate adversarial samples | The real neuron model used for inference |
|---|---|---|---|
| Scenario 1 | Yes | Neural oscillation | Neural oscillation |
| Scenario 2 | | Alternative neural oscillation | Alternative neural oscillation |
| Scenario 3 | No | LIF | Neural oscillation |
| Scenario 4 | | Alternative neural oscillation | Neural oscillation |
| Scenario 5 | | Neural oscillation | Alternative neural oscillation |

**Table 3** Top-1 classification accuracy (%) under the scenario 1 attack

| Models/Datasets | VGG-16/CIFAR-10 | | ResNet-18/CIFAR-10 | | ResNet-18/CIFAR-100 | |
|---|---|---|---|---|---|---|
| | benchmark | ours | benchmark | ours | benchmark | ours |
| Clean | 88.6 | 88.59 | 92.2 | 92.59 | 65.7 | 67.58 |
| FGSM | 14.2 | 30.9 | 35.5 | 44.9 | 12.6 | 17.2 |
| PGD-5 | 3 | 14.7 | 5.8 | 21.7 | 1.3 | 4.3 |
| BIM-5 | 1.9 | 14.3 | 5.9 | 21.4 | 1.4 | 4.5 |
| MIM-5 | 2.8 | 14.9 | 8.8 | 22.3 | 1.6 | 3.1 |

**Table 4** Top-1 classification accuracy (%) under the scenario 2 attack

| Models/Datasets | VGG-16/CIFAR-10 | | ResNet-18/CIFAR-10 | | ResNet-18/CIFAR-100 | |
|---|---|---|---|---|---|---|
| | benchmark | ours | benchmark | ours | benchmark | ours |
| Clean | 88.6 | 87.38 | 92.2 | 92.35 | 67.04 | 67.05 |
| FGSM | 14.2 | 52.2 | 35.5 | 66.9 | 12.6 | 45.1 |
| PGD-5 | 3 | 28.1 | 5.8 | 57.1 | 1.3 | 39.3 |
| BIM-5 | 1.9 | 31.6 | 5.9 | 59 | 1.4 | 39.8 |
| MIM-5 | 2.8 | 30.8 | 8.8 | 53.4 | 1.6 | 33.2 |

the effectiveness of this defense method based on the alternative neural oscillation in Section 4.

## 4. Results

We train SNN with the LIF model as the benchmark model to facilitate the comparison of the validation of our approach. The maximum perturbation size ε in all attacks is 8/255. For iterative attacks, for example, PGD-$i$, $i$ indicates the number of iterative steps. All attacks are non-target attacks.

### 4.1. Robustness analysis of neural oscillation model

We first test the robustness of the neural oscillation model against adversarial attacks. As shown in scenario 1 of Table 2, when attackers are fully aware of the structure, parameters, and form of the neural oscillation model, they use adversarial samples to attack networks. The results are presented in Table 3. Neural oscillation neuron always performs better than LIF neuron in terms of robustness in all three models/datasets under four attacks. The reason is that we introduce randomization. When noise is added to the neurons, this interferes with the validity of the perturbations superimposed in the input images. As illustrated in Figure 2(a), when noise $\gamma_i$ is different values, the gradient curve changes accordingly. This causes gradient value blurring and reduces the effectiveness of the generation of adversarial samples. Besides, $LeakyReLU$ leads gradients in specific sections to be small negative values, which makes it possible to cause the gradient's direction to blur. For example, in Figure 2(b), when H = 0.18 (green dashed line), the gradient of neuron with $\gamma_i = 0.8$ is positive, while the gradient of neuron with $\gamma_i = -0.2$ is negative. Thus, this randomness includes not only randomization in the value of the gradient but also the direction of the gradient.

### 4.2. Robustness analysis of alternative neural oscillation model

We then investigate the performance of alternative neural oscillation model with respect to robustness. As indicated in scenario 2 of Table 2, when attackers are fully aware of the structure, parameters, and form of the alternative neural oscillation model, they use adversarial samples to attack networks. The experimental results in Table 4 suggest that the alternative neural oscillation neurons have better robustness than LIF neurons under different adversarial attacks. And they are also more robust compared to neural oscillation neurons. The reason is that the alternative neural oscillation neuron replaces the random noise as a function of $H(t)$, which leads gradients $\frac{\partial S'(t)}{\partial H(t)}$ to be too steep (see Figure 2(a)). The steep gradients make the network back propagation optimization parameters unstable (gradient vanishing and exploding) and reduce the effectiveness of adversarial attacks. We have placed extra experimental results and argued this conclusion in the supplementary material.

### 4.3. Validation of defense

Scenarios 1 and 2 are both white-box attacks, where attackers are fully aware of all information about the network. Sometimes attackers only know part of the network, i.e. a grey-box attack. In this section, we test the validity of our defense strategy of masking real neurons with false neurons, thus tricking the attacker into generating attack samples with bias and reducing the efficiency of the attack. We consider three scenarios which are summarized in Table 2:

**Scenario 3:** Attackers are aware of the structure and parameters but do not know the form of neural model. Attackers use the LIF model to generate adversarial samples; however, neural oscillation is the real neural model for inference.

**Scenarios 4 and 5:** Attackers are aware of structure, parameters and know either the form of neural oscillation or the form of alternative neural oscillation. Attackers use the known 'fake' neurons to generate adversarial samples. The other neural model is as the real neuron of network for inference.

For scenario 3, Table 5 shows the top-1 accuracy of both benchmark and our method. The result demonstrates that our defense performance for both single-step attacks and iterative attacks on all three models is significantly better than SNNs. Since attackers do not know the specific expression of the neuron, this causes a significant decrease in the attack efficiency of the generated sample perturbations. For scenario 4, similar results to scenario 3 in Table 6 are observed: our method performs much better than benchmark against both single-step attack and iterative attack in all models and datasets. The results indicate that gradients of alternative neural oscillation neurons lead to a decreased attack success rate by masking original training gradients. For scenario 5, our method is still more robust to adversarial samples than benchmark in Table 7. However, it is worth noting that the defense of scenario 5 is much weaker when comparing the defense ability of scenario 4. When attackers know the expression of neural oscillation neuron, even though we replace them with the alternative neurons for inference, attackers can still generate effective adversarial samples to attack our networks. In other words, when attackers use alternative neural oscillation neuron to generate adversarial samples, the neuron, being very 'weak', is able to attenuate the attack capability of adversarial samples, thus indirectly enhancing the robustness of networks.

**Table 5  Top-1 classification accuracy (%) under the scenario 3 attack**

| Models/Datasets | VGG-16/CIFAR-10 | | ResNet-18/CIFAR-10 | | ResNet-18/CIFAR-100 | |
|---|---|---|---|---|---|---|
| | benchmark | ours | benchmark | ours | benchmark | ours |
| Clean | 88.6 | 88.59 | 92.2 | 92.59 | 67.04 | 67.58 |
| FGSM | 14.2 | 40.1 | 35.5 | 71.9 | 12.6 | 47.7 |
| PGD-5 | 3 | 40.8 | 5.8 | 79.3 | 1.3 | 56 |
| BIM-5 | 1.9 | 36.2 | 5.9 | 71.8 | 1.4 | 56.7 |
| MIM-5 | 2.8 | 26.7 | 8.8 | 67.6 | 1.6 | 49.4 |

**Table 6  Top-1 classification accuracy (%) under the scenario 4 attack**

| Models/Datasets | VGG-16/CIFAR-10 | | ResNet-18/CIFAR-10 | | ResNet-18/CIFAR-100 | |
|---|---|---|---|---|---|---|
| | benchmark | ours | benchmark | ours | benchmark | ours |
| Clean | 88.6 | 88.59 | 92.2 | 92.59 | 67.04 | 67.58 |
| FGSM | 14.2 | 56.5 | 35.5 | 69.5 | 12.6 | 43.7 |
| PGD-5 | 3 | 45 | 5.8 | 66.8 | 1.3 | 48.4 |
| BIM-5 | 1.9 | 41.4 | 5.9 | 68.5 | 1.4 | 51.4 |
| MIM-5 | 2.8 | 35.4 | 8.8 | 59 | 1.6 | 40.8 |

**Table 7  Top-1 classification accuracy (%) under the scenario 5 attack**

| Models/Datasets | VGG-16/CIFAR-10 | | ResNet-18/CIFAR-10 | | ResNet-18/CIFAR-100 | |
|---|---|---|---|---|---|---|
| | benchmark | ours | benchmark | ours | benchmark | ours |
| Clean | 88.6 | 87.38 | 92.2 | 92.35 | 67.04 | 67.05 |
| FGSM | 14.2 | 28.4 | 35.5 | 42.1 | 12.6 | 13.8 |
| PGD-5 | 3 | 12.1 | 5.8 | 19.2 | 1.3 | 4 |
| BIM-5 | 1.9 | 12.5 | 5.9 | 19.4 | 1.4 | 2.6 |
| MIM-5 | 2.8 | 13 | 8.8 | 20.2 | 1.6 | 2.4 |

**Table 8  White-box robustness (accuracy (%)) on CIFAR-10 using the ResNet-18 ($\epsilon=8/255$)**

| Defense | Clean | FGSM | PGD-20 |
|---|---|---|---|
| Standard (Madry et al. 2017) | 84.44 | 61.89 | 47.55 |
| MMA (Ding et al. 2018) | 84.76 | 62.08 | 48.33 |
| Dynamics (Wang et al. 2019) | 83.33 | 62.47 | 49.40 |
| TRADES (Zhang et al. 2019) | 82.90 | 62.82 | 50.25 |
| MART (Wang et al. 2019) | 83.07 | 65.65 | 55.57 |
| **Ours (scenario 4)** | **92.59** | **69.5** | **71.1** |

In fact, if we directly discard the neural oscillation neuron after training the model and replace it with the alternative neural oscillation neuron, then deploy the model in hardware devices, it is easy for attackers to be fooled by the 'fake' neuron.

Table 8 compares alternated neural oscillation with some advanced adversarial training defense methods in ANN. The adversarial training requires a large number of samples to retrain the model, and it is impractical to introduce all unknown attack samples into adversarial training, which would consume much time and computational resources, leading to the limitation of adversarial training. Our method only requires additional learning of a new oscillatory form through introducing only two parameters, which defends against most gradient-based adversarial attacks and is more efficient.

As neural oscillations are essential to many neural activities in the biological nervous system, SNN integrated with oscillation mechanism is more bio-plausible (In the supplementary we shows the spontaneous spike firing of our neural oscillation model, which is similar to the biological neural oscillation). Various mechanisms of the biological neural system provide a basis for optimizing the SNN, while these mechanisms integrated into the SNN also help us to better understand the biological neural system.

## 5. Discussion and conclusion

In this paper, we integrate brain-inspired neural oscillation into the SNN neural model and propose the neural oscillation neural model and alternative neural oscillation neural model. We verified that both neural models have better robustness than the LIF neuron. And we also use alternative neural oscillation neuron as the 'fake' neuron to defend against various gradient-based attacks. The experiments illustrate that our defense method can effectively resist both single-step attack and iterative attack.

Our method belongs to the class of methods that introduce randomization to enhance network robustness, but it is very different from the randomization currently used in ANNs. While previous literatures, such as (Liu et al. 2018), usually use random perturbations to disturb the generated samples, our method only introduces randomization over training and it is replaced by fitting a specific function during inference, which causes instability of the gradient. Therefore, our trained model has no randomization after training process, and the advantage of this approach is that when the attacker is fully aware of the neuronal model, the previous defense method only needs to remove the randomization to achieve an effective attack model, while our method cannot effectively attack the model after removing the fitting function. The attacker must work harder to find the original noise distribution in order to attack the model effectively. Thus, our method is more deceptive, which will make it difficult for the attacker to detect anomalies in the network.

There are still some limitations to our method. Since most of the current attacks are gradient-based attacks, our defense method was originally developed based on gradient-based attacks, so we did not test its effectiveness on other attack types. This part could be further explored in the future. Certainly, the neural oscillation model only partially mimics the form of biological neural oscillation; thus, further research might be conducted to integrate more complex neural oscillation forms in SNNs.

## Supplementary

### A.1 Parameter values for reproducibility

Table A.1.1 shows the noise range and parameters c and d. In our work, we picked [a,b] in the range [-0.2,0.8]. The noise is generally selected not to exceed the threshold $V_{th}$ (otherwise it may lead to a reduction of accuracy) and is mainly concentrated between $V_{th}$ and the $V_{reset}$. There is no mandatory range size for the selection of these hyperparameters, and we actually tried different ranges and were able to obtain similar defensive effects. For noise type, in the main paper we use the random uniform noise, we have also tried the Gaussian noise, and it can also be fitted by different equations to generate the alternative neural oscillation neuron.

Table A.1.1 Noise range [a,b] and values of $c$ and $d$ of alternative neural oscillation model

| Model/Dataset | $[a,b]$ | $c$ | $d$ |
|---|---|---|---|
| VGG-16/CIFAR-10 | [-0.2,0.8] | -0.1441 | -0.1762 |
| ResNet-18/CIFAR-10 | | -0.1019 | -0.2221 |
| ResNet-18/CIFAR-100 | | -0.1564 | -0.1687 |

### A.2 Neuron performance testing

Figure A.2.1 depicts the accuracy of the SNNs on the corresponding architecture/dataset when using either neural oscillation neuron or alternative neural oscillation neuron. In Figure (a), the VGG-16 network composed of alternative neural oscillation neurons cannot even be optimized. In Figure (b), the alternative neural oscillation neuron significantly decreases the network optimization's speed and accruacy. Thus, the results indicate that alternative neural oscillatory neurons have worse performance than neural oscillatory neurons, which tend to lead to instability of the gradient (e.g. gradient vanishing or explosion), and hence make the network less capable of optimization. When an attacker generates the adversarial samples with such neurons, the adversarial samples also become less powerful.

### A.3 Function selection of alternative neural oscillation

In the main paper, we use the $sin(H(t) + c) + d$ to fit the random uniform noise item $\gamma(t)$. In practice, the function can be of many different forms, such as equations (1)-(4). These equations all fit the random noise term $\gamma(t)$ well after training parameters $c$ and $d$.

$$F1 = sin(H(t) + c_1) + d_1 \quad (1)$$

$$F2 = x * sin(H(t) + c_2) + d_2 \quad (2)$$

$$F3 = e^{(x+c_3)} + d_3 \quad (3)$$

$$F4 = \frac{1}{1+e^{-c_4 x}} + d_4 \quad (4)$$

We draw the *F* curves in Fig.A.3.1(a) and the corresponding gradient curves Fig.A.3.1(b) when different *F* is chosen, respectively. Table A.3.1 compares the effectiveness of the defense when using different *F*. As we see, *F2* provides the best defense against all kinds of attacks, while *F4* is the worst. And the gradient curve of *F2* shown in Fig.A.3.1(b) is the steepest, when the gradient curve of *F4* is the flattest.

These results argue our view that the steep gradient causes instability in the network, which weakens the effectiveness of generated adversarial samples.

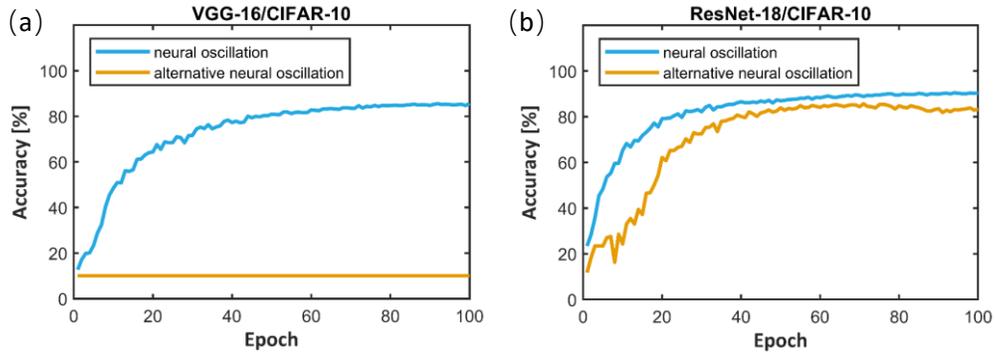

Fig.A.2.1 Accuracy on (a)VGG-16/CIFAR-10 (b)ResNet-18/CIFAR-10 when using neural oscillation neuron (blue curve) and alternative neural oscillation neuron (orange curve)

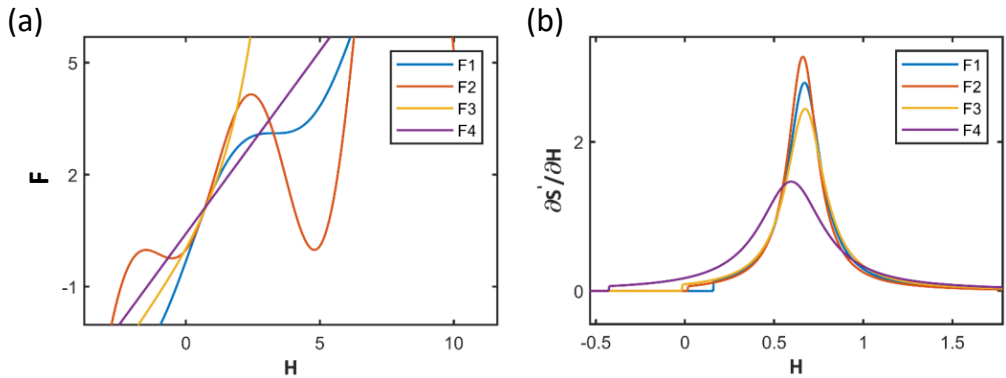

Fig.A.3.1 (a) Curve of function $F$ to fit the noise item on VGG-16/CIFAR-10. (b) Gradient curve $\frac{\partial S'(t)}{\partial H(t)}$ when using different $F$.

Table A.3.1    Top-1 classification accuracy (%) under the scenario 4 attack

| Models/Datasets | VGG-16/CIFAR-10 | | | | |
|---|---|---|---|---|---|
| | benchmark | *F1* | *F2* | *F3* | *F4* |
| Clean | 88.6 | 87.38 | 86.12 | 87.7 | 86.64 |
| FGSM | 14.2 | 56.5 | 68.9 | 50.7 | 36.8 |
| PGD-5 | 3 | 45 | 68.8 | 34.6 | 22.7 |
| BIM-5 | 1.9 | 41.4 | 69.3 | 32.9 | 15 |
| MIM-5 | 2.8 | 35.4 | 60.4 | 31.4 | 20.6 |

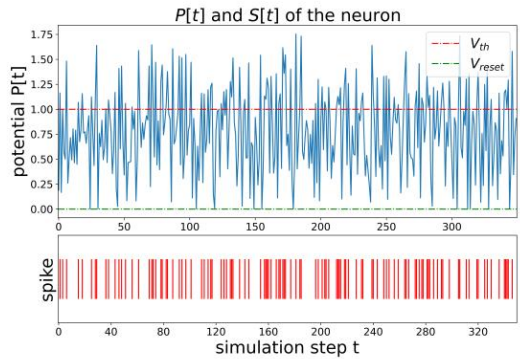

Fig.A.4.1 Spontaneous spike firing of neural oscillation neuron.

## A.4 Firing property of neural oscillation neuron

Neural oscillation arises from the spontaneous spike firing behavior of biological neurons. This spontaneous behavior is not influenced by external stimuli. In our proposed neural oscillation neural model, the inclusion of random noise allows the neuron to generate spontaneous spike firing in the absence of input, as shown in Figure A.4.1. This property makes our model more bio-plausible.